%% file: acl2021.tex
\documentclass[11pt,a4paper]{article}
\usepackage{authblk}
\usepackage[hyperref]{acl2021}
\usepackage{times}
\usepackage{latexsym}

\usepackage{microtype}
\usepackage{booktabs}
\usepackage{multirow}
\usepackage{graphicx}
\usepackage{subcaption}
\usepackage{xspace}
\usepackage{multirow}
\usepackage{lipsum}

\newcommand{\BLEC}{\textsc{Blec}\xspace}
\newcommand{\snowball}{\textsc{Snowball}\xspace}
\makeatletter
\newcommand\email[2][]%
   {\newaffiltrue\let\AB@blk@and\AB@pand
      \if\relax#1\relax\def\AB@note{\AB@thenote}\else\def\AB@note{\relax}%
        \setcounter{Maxaffil}{0}\fi
      \begingroup
        \let\protect\@unexpandable@protect
        \def\thanks{\protect\thanks}\def\footnote{\protect\footnote}%
        \@temptokena=\expandafter{\AB@authors}%
        {\def\\{\protect\\\protect\Affilfont}\xdef\AB@temp{#2}}%
         \xdef\AB@authors{\the\@temptokena\AB@las\AB@au@str
         \protect\\[\affilsep]\protect\Affilfont\AB@temp}%
         \gdef\AB@las{}\gdef\AB@au@str{}%
        {\def\\{, \ignorespaces}\xdef\AB@temp{#2}}%
        \@temptokena=\expandafter{\AB@affillist}%
        \xdef\AB@affillist{\the\@temptokena \AB@affilsep
          \AB@affilnote{}\protect\Affilfont\AB@temp}%
      \endgroup
       \let\AB@affilsep\AB@affilsepx
}               
\newcommand\blfootnote[1]{%
\begingroup
\renewcommand\thefootnote{}\footnote{#1}%
\addtocounter{footnote}{-1}%
\endgroup
}

\makeatother
\aclfinalcopy 

\title{Logic-Consistency Text Generation from Semantic Parses}
\author[1*]{\textbf{Chang Shu}}
\author[2*]{\textbf{Yusen Zhang}}
\author[3]{\textbf{Xiangyu Dong}}
\author[4]{\textbf{Peng Shi}}
\author[5]{\textbf{Tao Yu}}
\author[2]{\textbf{Rui Zhang}}

\affil[1]{School of Informatics, University of Edinburgh}
\affil[2]{Department of Computer Science and Engineering, Penn State University}
\affil[3]{School of Computer Science and Engineering, Beihang University}
\affil[4]{David R. Cheriton School of Computer Science, University of Waterloo}
\affil[5]{Department of Computer Science, The University of Hong Kong}

\email{\texttt{s1783039@ed.ac.uk, yfz5488@psu.edu, dxy912434027@gmail.com}}
\email{\texttt{peng.shi@uwaterloo.ca, rmz5227@psu.edu}}

\date{}

\begin{document}
\maketitle
\blfootnote{*Equal Contribution}

\begin{abstract}
Text generation from semantic parses is to generate textual descriptions for formal representation inputs such as logic forms and SQL queries. This is challenging due to two reasons: (1) the complex and intensive inner logic with the data scarcity constraint, (2) the lack of automatic evaluation metrics for logic consistency. To address these two challenges, this paper first proposes \snowball, a framework for logic consistent text generation from semantic parses that employs an iterative training procedure by recursively augmenting the training set with quality control. Second, we propose a novel automatic metric, \BLEC, for evaluating the logical consistency between the semantic parses and generated texts. The experimental results on two benchmark datasets, \textit{Logic2Text} and \textit{Spider}, demonstrate the \snowball framework enhances the logic consistency on both \BLEC and human evaluation. Furthermore, our statistical analysis reveals that \BLEC is more logically consistent with human evaluation than general-purpose automatic metrics including BLEU, ROUGE and, BLEURT. Our data and code are available at \url{https://github.com/Ciaranshu/relogic}.
\end{abstract}

\section{Introduction}

\begin{figure}[ht]
\centering
  \includegraphics[width=\linewidth]{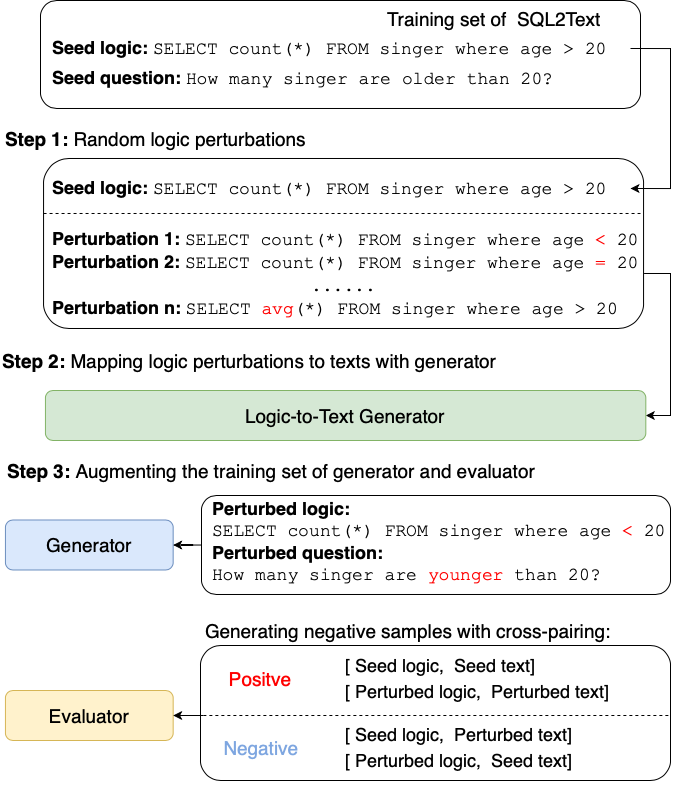}
  \caption{Our data augmentation procedure for the generator and evaluator in the \snowball framework.
}
  \label{data_aug}
\end{figure}

\begin{figure*}[ht]
  \includegraphics[width=\textwidth]{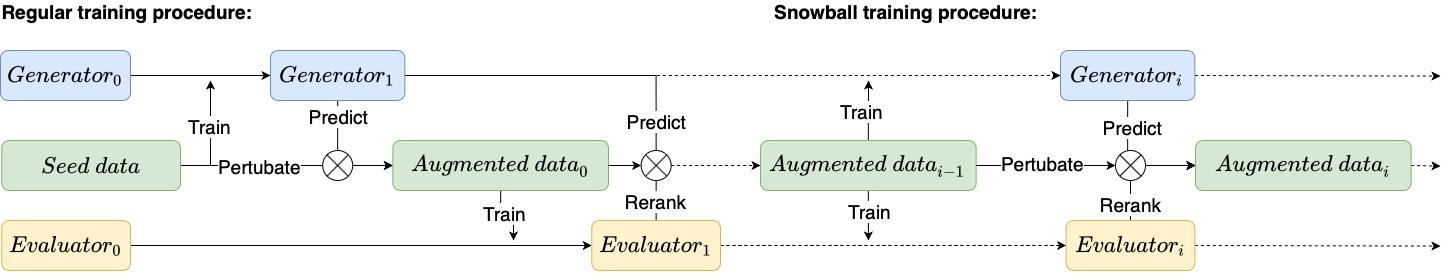}
  \caption{
  Our \snowball framework employs an iterative training procedure over a generator and evaluator through data augmentation.
}
  \label{snowball}
\end{figure*}

Natural language generation (NLG) from semantic parses is to generate the text description for the formal representation input such as logical forms, AMR, and SQL queries. It has drawn widespread attention because of its substantial contributions to the interpretability and usability of the latest natural language interfaces~\cite{DBLP:journals/jair/GattK18,chen-etal-2020-logic2text, DBLP:journals/ese/HuLXLJ20,mishra-etal-2019-storytelling,yu-etal-2019-cosql,DBLP:conf/www/NgomoBULG13,wang-etal-2018-describing,gardent-etal-2017-webnlg,wang-etal-2020-towards,wang-2019-revisiting,DBLP:conf/icde/KoutrikaSI10}. Recently, pre-trained large-scale language models like BERT~\cite{devlin2018bert}, T5~\cite{DBLP:journals/jmlr/RaffelSRLNMZLL20}, and GPT-3~\cite{brown2020language} have raised the ability to generate natural language from formal texts to a promising level of fluency and coherence. 

However, NLG from semantic parses still has suffered from two crucial challenges: (1) the data scarcity constraint due to the bias on certain types of logic forms or expensive labeling work~\cite{iyer2017learning,yaghmazadeh2017sqlizer}, which potentially leads to the unsatisfied fidelity of remaining the complex and intensive inner logic in the generated text based on our empirical research; (2) The general-purpose automatic metrics~\cite{novikova-etal-2017-need} such as BLEU~\cite{papineni-etal-2002-bleu}, ROUGE~\cite{lin-2004-rouge} and BLEURT~\cite{sellam-etal-2020-bleurt} are not ideal for explicitly measuring the logic consistency~\cite{wang2020faithful, DBLP:journals/corr/abs-2004-06577}, because they tend to evenly weight each word in the generated text without fully attending on the fatal logical keywords.

To address these two critical problems, we propose the \snowball framework for high-fidelity text generation from semantic parses and the \BLEC automatic evaluation metric for logic consistency:\\
\textbf{Snowball Framework.} Our \snowball framework, as illustrated in Figure~\ref{snowball}, trains two modules to ensure high-fidelity text generation: (1) a generator that maps the logical form to its textual description, and (2) an evaluator that indicates the logic consistent score of each pair of logical form and textual sentence. Rather than training the generator and evaluator independently, \snowball performs iterative training on the generator and the evaluator.
To deal with the data scarcity issue, we propose a data augmentation procedure to cover valid logic variations with diverse natural language expressions to improve generalizability.
To this end, during each iteration, various unseen logic pairs could be automatically generated with rule-based enumerated logic forms and their corresponding text predicted by the generator. The evaluator is then used to filter out the high reliable augmented logic pairs for the next training iteration.\\
\textbf{\BLEC Metric.}
To evaluate the logic consistency of the text generated by the model, we propose a rule-based automatic evaluation metric called Bidirectional Logic Evaluation of Consistency, or \BLEC. It takes the logical form and the generated corresponding natural language text as input, then outputs a label indicating if they represent consistent logic. Compared with the neural network evaluator, \BLEC can be easily deployed to different datasets, as long as the parser (i.e., the grammar of the logical form) is given.

In our experiments, we exam the effectiveness of our proposed approaches on the benchmark datasets of NLG from semantic parses derived from existed Text-to-SQL dataset \textit{Spider}  \cite{DBLP:conf/emnlp/YuZYYWLMLYRZR18} and Table-to-Text dataset \textit{Logic2Text} \citet{chen-etal-2020-logic2text}. Our analysis shows that our \BLEC metric has a substantially positive Pearson score with human annotations, demonstrating better logic consistency than other automatic metrics. The \BLEC result shows that the \snowball framework leads to accordant enhancement in logic consistency on two datasets compared to the single-pass training method based on BART~\cite{DBLP:conf/acl/LewisLGGMLSZ20}.

Our key contributions are summarized into three-folds: (1) We propose a simple but effective training framework \snowball that strengthens the logic faithfulness of generated text by covering diverse logic variations. (2) We propose a new logic evaluation metric \BLEC that accurately measures the logical consistency with a refined keyword matching mechanism. (3) Our experiment results demonstrate that \snowball at most increases the \BLEC from 10.1\% on SQL-to-Text and 1.2\% on Logic-to-Text tasks compared to the baseline. Moreover, our statistical analysis reveals that \BLEC achieves a +0.66 Pearson correlation coefficient compared with human labels, serving as a much better automatic evaluation metric than not only the traditional BLEU and ROUGE metrics, but the latest BLEURT metrics.

\section{Related Work}

\subsection{Parses-to-Text}
The source of data-to-text (D2T) datasets is mostly a flat ontology structure, like E2E\cite{novikova-etal-2017-e2e}, LogicNLP\cite{chen-etal-2020-logical}, RotoWire\cite{wiseman-etal-2017-challenges}, and ToTTo\cite{parikh-etal-2020-totto}, which is not powerful enough to encode rich semantic relationships in the ontology. Second, some datasets, such as WebNLG\cite{gardent-etal-2017-webnlg}, E2E, and RotoWire, have a limited number of domains. E2E is on the restaurant domain, and RotoWire is on the basketball domain. Moreover, some of them only have loose alignments between input and sentence, e.g., RotoWire.

Generating the natural language descriptions for the logic forms or parses as a sub-task of D2T, has been studied in various datasets and tasks, such as GCC grammar to text~\cite{white2006efficient}, and UCC grammar to text~\cite{gardent1990generating}. There are a lot of works that leverage the neural networks to conduct the generation on various tasks, for example, generating natural language from AMR~\cite{song-etal-2018-graph,ribeiro2019enhancing,damonte2019structural}, logic forms~\cite{chen-etal-2020-logic2text}, as well as SQL parses~\cite{xu2018sql,ngonga2013sorry,koutrika2010explaining}. However, different from these works, our work focuses on the logic consistency generation from parses. So we will mainly discuss and evaluate the model based on the logic between parses and questions. 

\subsection{High-fidelity Text Generation}
As for the end-to-end neural-based text generation models, collaborating the auxiliary task during model training is an intuitive method that introduces the logic regulation to the models. For instance, the fidelity classification task proposed by \citet{DBLP:journals/corr/abs-2004-06577}, the auxiliary span extraction tasks by \citet{kryscinski-etal-2020-evaluating},the table-text optimal-transport matching and embedding similarity losses by \citet{wang2020faithful} and the content matching task presented by \citet{parikh-etal-2020-totto} are proved to be effective. Nevertheless, to the best of our knowledge, we are the first to bridge the training procedure of evaluator and generator together with the iterative training framework snowball. Furthermore, we attempt to construct a new automatic metric and a new dataset dedicated to evaluating the logic consistency of text generation. The concentration of our work differs from the related high-fidelity text generation work \cite{chen-etal-2020-logic2text,chan-etal-2019-stick,nie-etal-2018-operation,tian2019sticking,wang-etal-2020-towards}, by attempting to present the panorama of the challenges of logic-consistent text generation instead of focusing on the model-wised modifications.

\section{Snowball Framework}

The \snowball framework addresses the challenge of the complex and intensive inner logic with data sparsity constraint for the high-fidelity text-generation from semantic parses. As illustrated in Figure~\ref{snowball}, \snowball assures the logic consistency with three bases: (1) Iterative training procedure synergistically enhances the generator and evaluator in the adversarial fashion; (2) Data augmentation based on rule-based logic perturbations and neural-based text generation covering diverse unseen logic variations for iterative training; (3) Structure-aware encoding boost the sensibility of the encoder on mild logic shift.

\subsection{Iterative Training}
Rather than training the generator and evaluator independently, \snowball performs training on the generator and the evaluator iteratively. As demonstrated in Figure~\ref{snowball}, the prerequisite of the snowball training procedure is the regular training procedure: (1) the $Generator_0$ is trained on the benchmark NLG datasets with the normal end-to-end approach into trained $Generator_1$; (2) meanwhile, the logic forms in the seed data are converted into variations with given rules, then the $Generator_1$ predicts the text for each mutated logic forms to be a completed logic pair; (3) The initial $Evaluator_0$ is then trained on those augmented logic pairs.

Then, during the \snowball procedure, the generator and evaluator are collaboratively improved through several training iterations, and during each iteration, a three-step adversarial interaction would be conducted between the generator and evaluator: \textbf{Step 1:} The trained $Evaluator_{i-1}$ could be used to rerank the beam search results given by the decoder of the generator, consequently leading to increased quality of the augmented logic pairs, $Augmented\hspace{1mm} data_{i-1}$; \textbf{Step 2:} The $Generator_i$ is capable to better retain the logic consistency by training on the $Augmented\hspace{1mm}  data_{i-1}$ which contains more unseen logic variations uncovered in the seed data; \textbf{Step 3:} The enhanced $Generator_i$ predicts the increasingly realistic-like perturbed sentences from the perturbed logical forms, which brings more challenging negative samples to the training set of the $Evaluator_i$. The data augmentation in the first step would be further described in Section~\ref{data_aug_sec}.

To be specific, our generator and evaluator in \snowball are described as follows.
\paragraph{Generator}
The generator maps the logical form to the corresponding natural language sentences.
We choose the pre-trained BART model~\cite{DBLP:conf/acl/LewisLGGMLSZ20} following the standard transformer architecture~\cite{DBLP:conf/nips/VaswaniSPUJGKP17}, which contains the encoder and decoder architecture as the denoising autoencoder pre-trained on the task of corrupted text reconstruction. The input of the encoder is the structure-aware representation of the logic forms (Section \ref{struct}), while the target output of the decoder is the aligned textual description for the input parses.

\paragraph{Evaluator}
An evaluator indicates the logic consistent score of pairs of logical forms and textual sentences, which is vital for assessing the performance of the logic-focused text generator. In contrast to other text generation tasks, generating sentences from logical forms especially requires the evaluator to be reasonably sensitive to the subtle logic shifts of the model predictions. For instance, deleting negation words such as `not' is fatal for our task by significantly compromising the logic consistency. Therefore, we exploit a binary classification architecture similar to the BART-based natural language inference model \cite{DBLP:conf/acl/LewisLGGMLSZ20} as our evaluator to compute the consistency between the pairs of logical form and text $[L, Q]$. The input of the encoder is the concatenation of the L and Q appended an $[EOS]$ token, and the logic scores $\gamma$ are computed as:
\begin{equation}
\gamma = \sigma(\omega([h_{d_1},h_{d_2},h_{d_3}...]))
\end{equation}
where $h_{d_n}$ denotes the last hidden states of the decoder, $\omega$ denotes the max-pooling layer, and $\sigma$ is the sigmoid activation function.

\subsection{Data Augmentation}\label{data_aug_sec}

\label{struct}
\begin{figure*}[ht]
\centering
  \includegraphics[width=\textwidth]{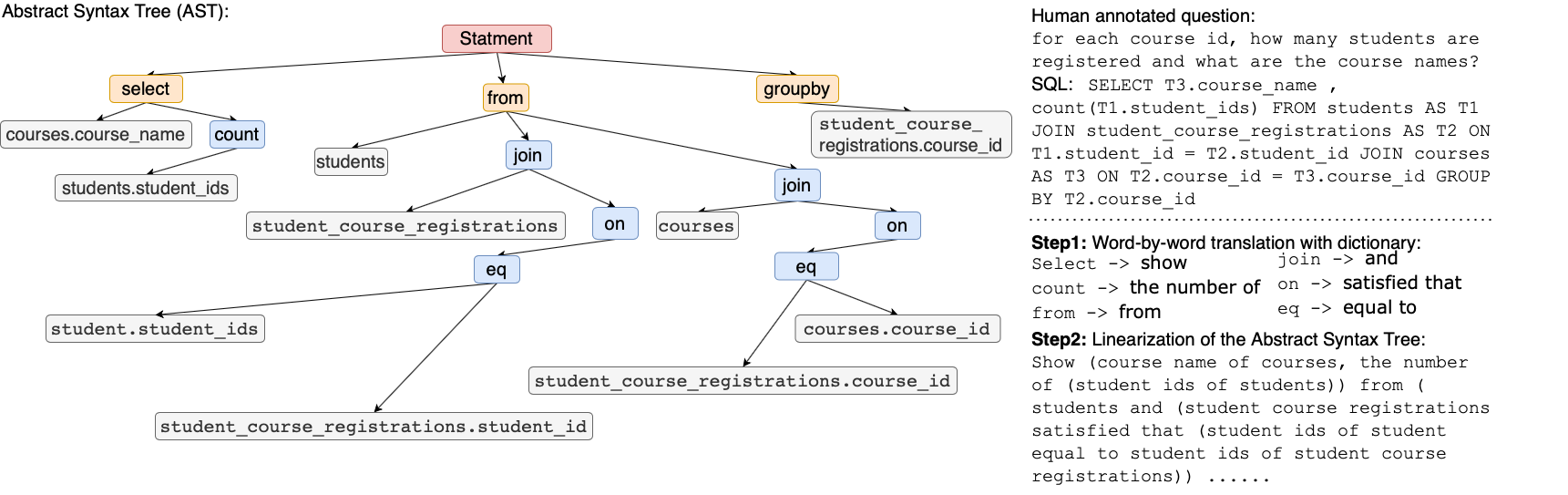}
  \caption{The example of word-by-word translation and linearization of structure-aware encoding.}
  \label{encoder}
\end{figure*}
As the labeled training data for both the generator and evaluator is extremely limited, we propose a data augmentation procedure to enlarge the training set by covering variations of logic forms paired with diverse natural language expressions to improve the generalizability. To be specific, our data augmentation consists of three steps as depicted in Figure~\ref{data_aug} from a seed dataset with human annotation:
\paragraph{Step 1: Logic perturbation}
Instead of modifying the natural language sentences, we choose to corrupt the logic consistency by perturbing logical forms mainly because of two reasons: (1) The regular structures of logical forms guarantee the procedure of the logical corruption to be comparatively controllable; (2) The perturbed logical forms could be easily validated with the corresponding parser and grammar checker. The perturbations of each given logical form could be enumerated exhaustively according to hand-tuned rules to cover the following logic inconsistencies:
\begin{itemize}
\item \textbf{Logic shift:} The logic shift indicates that the generated text logically distinct from the input logical forms, such as turning the assertive sentences into negative sentences. This could be attributed to the perturbations of aggregators, operators, logic conjunction, etc.
\item \textbf{Phrase and number changes:} The phrase changes mean that the generated sentence modifies the appointed phrase from the logical forms, while the number changes are that the numerical values in the logical forms are perturbed. 
\item \textbf{Entity insertion, deletion and swapping:} Perturbations of entities is a common drawback that most natural language generation models suffer. This includes the phenomenon that the predicted sentences neglect the entities mentioned in the logical forms, insert unrelated entities to the logical form, or mislay them.
\end{itemize}
\paragraph{Step 2: Inference from perturbed logic}
After logic perturbation, the generator could be exploited as the artificial annotator to generate the corresponding sentence for each logical form in a semi-supervised manner. Compared to the rule-based or template-based method, the recent pre-trained seq-to-seq models empirically generate the natural language sentences with better fluency and coherency. Though this method could easily create a considerable amount of labeled data meanwhile avoid the expensive human annotation, what can not be ignored is that the model-based generator naturally would introduce unexpected noise during augmentation. Therefore, the quality control for the data augmentation is one of the most crucial cornerstones for a satisfactory result.
\paragraph{Step 3: Dataset composition}
As shown in Figure~\ref{data_aug}, the example in the seed dataset is denoted as [Seed logic, Seed text], and the augmented examples are denoted as [Perturbed logic, Perturbed text]. Intuitively, we may take the augmented [Perturbed logic, Perturbed text] to be not only the training example for the generator but also the positive sample for the evaluator, while crossover pairs [Seed logic, Perturbed text] and [Perturbed logic, Seed text] would be suitable negative samples for the evaluator.

\subsection{Structure-aware Encoding}

The logical forms normally have equivalent structured representations to precisely express the complex relations between a set of objects. For instance, the executable codes written in Python or SQL could be parsed into abstract syntax tree(AST) \cite{DBLP:journals/cl/Noonan85} denoting the mutual relations among occurred constructs in the source code, while the knowledge bases may be converted into knowledge graphs that depict the relations between entities with directed edges. Compared to the plain text inputs, the structure-aware encoding capturing not only the sequential information from texts but also the internal logic from structural representations recently has been proved to be more effective in several Graph-to-Text tasks \cite{song-etal-2018-graph, DBLP:journals/corr/abs-2007-08426}. To make full use of the intrinsic knowledge of the pre-trained BART model, we follow the similar approach proposed by \citet{DBLP:journals/corr/abs-2007-08426} to linearize the structural representations of the SQL queries and logical forms respectively (Figure \ref{encoder}). Furthermore, the logical forms from different domains or datasets may vary in keywords, so normalizing them into a unified form would bridge the gaps between different logic NLG datasets and then increase the generalization ability of our framework. Hence, the logical forms would be firstly word-by-word translated into the unified intermediate semi-textual forms according to a manually annotated dictionary. Then the parenthesis is inserted into the semi-textual forms to denote the hierarchy of the correlated structured representations such as ASTs.

\section{\BLEC for Logic Consistency Evaluation}
\begin{figure}[t]
    \centering
    \includegraphics[width=\linewidth]{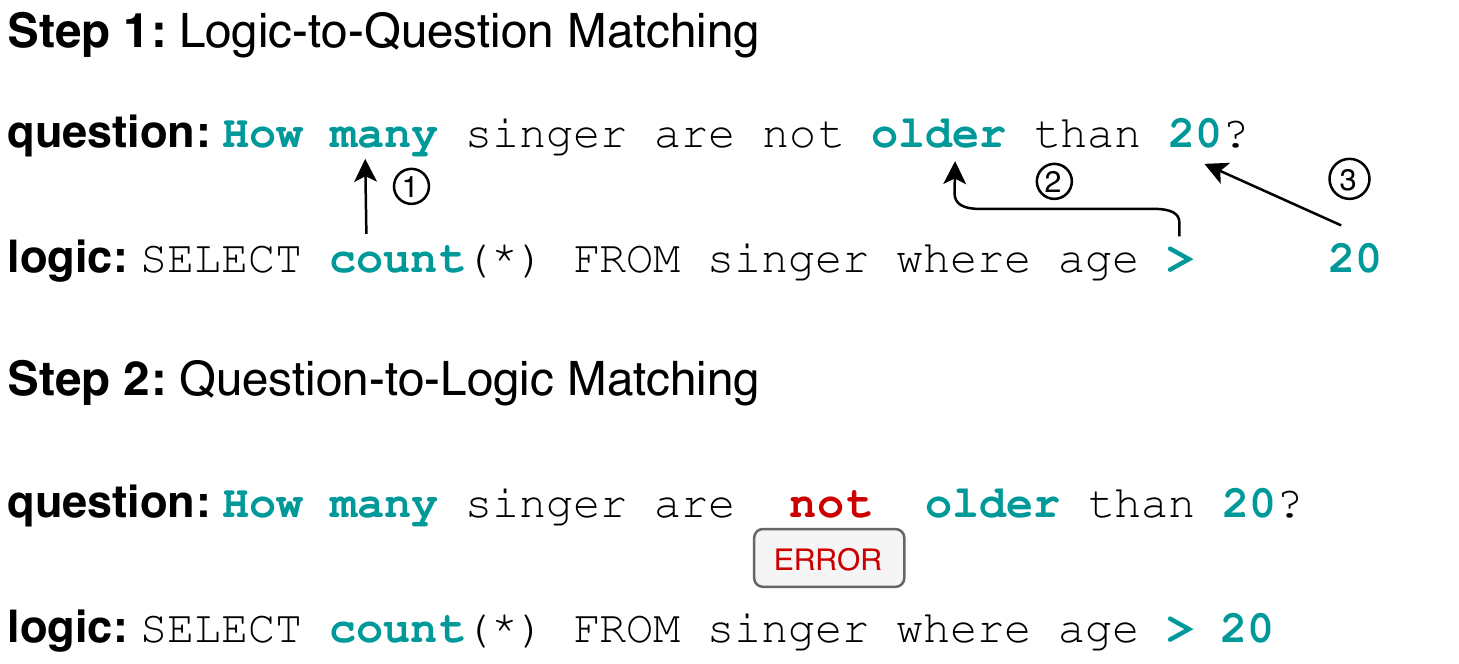}
    \caption{A sample of \BLEC. The words marked in green are the matched tokens while the words marked in red are the tokens with no match.}
    \label{fig:BLEC}
\end{figure}

Because the general-purpose automatic metrics such as BLEU, ROUGE, and BLEURT are not ideal for explicitly measuring the logic consistency, we propose \BLEC, a new rule-based automatic evaluation metric called Bidirectional Logic Evaluation of Consistency.
We apply a bidirectional evaluation to determine the logical consistency of pairs of logical forms and questions. The intuition behind this metric is that some key tokens such as number, operator, and keywords in the logical form should always be matched with some tokens that represent similar meanings in the question, and vice versa. An example is shown in Figure~\ref{fig:BLEC}, \BLEC first traverses the key tokens in the question, trying to find the tokens with the same meaning in the logic form to match them. Then, in step two, the sample is marked as inconsistent because there is one token with no match from the question to the logical form.

Formally, given a logical form $L = l_1,l_2,...,l_n$ containing $n$ word tokens and a questions $Q = q1,q2,...,q_m$ containing $m$ word tokens, the proposed evaluation metric performs token level matching on $l_i$ and $q_j$ to test the consistency.
To be specific, the matching procedure contains two steps, i.e. matching from $L$ to $Q$ as well as matching from $Q$ to $L$. In step one, each key token $l^{key}_i$ in $L$ tries to match with the tokens in $Q$. In step two, each key token $q^{key}_j$ in $Q$ tries to match with the tokens in $L$. If no tokens are found that could be matched with any key tokens in either step one or step two, the sample will be marked as negative, vice versa. The final score is the accuracy of all the samples:
\begin{equation}
\BLEC = \frac{\sum_{s\in S} match(s)}{|S|}
\end{equation}

Where $S$ denotes the dataset while $match(*)$ is the matching function with binary output, i.e. 1 for positive and 0 for negative.

Compared with the neural network evaluator requiring data-specific training, \BLEC can be easily deployed to different datasets. In our experiments, we demonstrate that \BLEC can be applied to two different datasets of text generation from two types of semantic parse input, and it shows a substantial agreement with human evaluation for evaluating logic consistency between the semantic parse input and the text output (Table~\ref{tab:pscore}).

\section{Experiment Settings}

\begin{table}[t!]
\small
\centering
\begin{tabular}{@{}llccc@{}}
\toprule
\multicolumn{2}{l}{\multirow{2}{*}{Dataset}} & \multicolumn{1}{c}{Train} & \multicolumn{1}{c}{Dev} & \multicolumn{1}{c}{Test}  \\
\multicolumn{2}{l}{}                         & \multicolumn{1}{c}{}      & \multicolumn{1}{c}{}    & \multicolumn{1}{c}{}           \\ \midrule
\multirow{2}{*}{SQL2Text}   & Generator & 5600 & 1400 & 1034  \\
                            & Evaluator & -    & 1142 & 1142  \\ \midrule
\multirow{2}{*}{Logic2Text} & Generator & 8566 & 1095 & 1092  \\
                            & Evaluator & -    & 1041 & 1041    \\ \bottomrule
\end{tabular}%
\caption{The statistics of the SQL2Text and Logic2Text dataset.}
\label{tab:dataset}
\end{table}

\subsection{Datasets}
Text generation from semantic parses has different forms depending on the input formal representation.
To demonstrate that our \snowball and \BLEC can be applied to different types of inputs, we study two tasks: (1) SQL2Text with the SQL query as the input and (2) Logic2Text with the logic forms as the input.

To this end, we make use of two existing publicly available datasets:
For SQL2Text, we use the Spider dataset \cite{DBLP:conf/emnlp/YuZYYWLMLYRZR18}, a complex cross-domain semantic parsing and text-to-SQL dataset. Generating natural language from formal languages with abundant logic representations could be regarded as the inverse semantics parsing process. Therefore, we reverse the input and output as a dataset for the text generation from SQL queries with complicated logic. As the test set of the Spider dataset remains undisclosed, 20\% of the original Spider training set is converted into a development set, and 80\% of the training set remains to be the training set, and the original development set is exploited as the test set for our SQL2Text task.
For Logic2Text, we use an existing Logic2Text dataset from \citet{chen-etal-2020-logic2text}. We pick the \textsc{sent} and \textsc{logic\_str} fields from the original Logic2Text to compose our own train data. We then change \textsc{sent} to \textsc{text}and change \textsc{logic\_set} to \textsc{logic} as our one keyword of each sample in the dictionary of our dataset.

In contrast, evaluating the logical consistency between logical form and text is closely related to the sequence classification tasks such as fact verification and natural language inference (NLI). According to the best of our knowledge, there is no existing dataset for evaluating the logical consistency between logical form and generated text. Therefore, we simplified the logic evaluation as a two-sequence binary classification problem and then construct the dataset with the development set and test set dedicated for our proposed evaluator. The dataset is constituted from the development and test set of Spider and Logic2text by three methods: (i) The [logical form, Text] pairs in the two datasets are regarded as positive samples; (ii) The human-labeled negative samples by intentionally introducing the logical inconsistency to the known [logical form, Text] pairs in the two datasets; (iii) The manually scored [logical form, Text] prediction given by the trained generator on the two datasets which contain both positive and negative samples.
As for the human-labeled negative samples, we attempt to cover the possible logic perturbations mentioned in section~\ref{data_aug_sec} with minimum modification to the original [logical form, Text] pairs. For example, a coincident pair [\texttt{SELECT avg(age) FROM dogs}, What is the \textbf{average} age of dogs?] would be corrupted into  [\texttt{SELECT avg(age) FROM dogs}, What is the \textbf{oldest} age of dogs?]. Table~\ref{tab:dataset} summarizes the statistics of each dataset for both generator and evaluator, respectively. 

\subsection{Baselines and Implementation Details}
The baselines for assessing the performance of \snowball framework are the attention-based LSTM machine translation model \cite{tao-etal-2019-effective}, and the single-pass trained models which are the models trained before performing \snowball iteration.  For instance, the \texttt{BART-large} generator trained in the second \snowball iteration would be compared to the identical \texttt{BART-large} generator in the zero \snowball iteration.
The hype-parameter settings of the models trained on \textit{SQL2Text} and \textit{Logic2Text}, mostly follow the default setting of BART model from Huggingface~\cite{DBLP:conf/acl/LewisLGGMLSZ20, wolf-etal-2020-transformers}. However, the learning rate of evaluator and tokenizer are slightly different, namely the learning rate of evaluator on \textit{SQL2Text} is 2e-5 for \texttt{BART-base} and is 5e-6 for \texttt{BART-large}, while the learning rate of evaluator on \textit{Logic2Text} is 1e-5 for both \texttt{BART-base} and \texttt{BART-large}.

\subsection{Multitask Learning}
Due to the lack of data of logic NLG, intuitively collaborative training on \textit{SQL2Text} and \textit{Logic2Text} dataset may prevent the models from bias fitting to their confined training data. Aside from the standard special separators used by BART tokenizer, we further introduce \textit{[SQL]} and \textit{[logic]} tokens to be the control codes to indicate if one sample is from \textit{SQL2Text} or \textit{Logic2Text} dataset, similar as \cite{DBLP:journals/corr/abs-1909-05858}. For each sample fed into the BART model, a corresponding control token is contacted in the front of the input logical form according to that sample source. Therefore, the distribution $p(Q_{SQL}|L_{SQL},[SQL])$ of the \textit{SQL2Text} models and $p(Q_{Logic}|L_{Logic}, [logic])$ of the \textit{Logic2Text} models could be learned respectively during the backpropagation that takes the control tokens into account, while training the generator and evaluator in the MTL fashion.

\subsection{Human Evaluation}
To evaluate if the sentence generated by the model is logically consistent, we randomly sample 90 questions from the test set of Spider and a test set of logic2text separately to form a human evaluation set. The samples of each setting will be divided into two parts and assigned to two different annotators. Each part contains 10 overlap and 40 non-overlap examples, which means one person has to label 50 samples for a setting. As for the human evaluation, the annotators label the [logical form, text] as True or False based on two criteria: (1) the logic consistency between logical form and text; (2) The grammaticality of the text. After labeling, we estimate the accuracy of the model predictions by computing the expectation of the true labels from 80 non-overlap data. To prove the consistency of the annotators, we use the 10 overlap data to calculate the cohen kappa score. Only if the kappa score is over 0.4 which implies that this estimated accuracy is valid, the results would be reported. In Table \ref{tab:generator}, we only human annotated the results given by the models trained without snowball iteration and trained with 4 snowball iterations. It demonstrates the correlation between human evaluation and BLEC metrics in these two time steps instead of directly evaluating the improvement of model performance.  

\section{Results and Analysis}

\subsection{Correlation Analysis on \BLEC}
To show that \BLEC is consistent with human judgment, we test the Pearson correlation between the \BLEC score and the human evaluation result. We also include ROUGE and BLEU for comparisons. Therefore, we apply these four automatic metrics (BLEU, ROUGE, BLEURT, \BLEC) to a human-labeled dataset and compare the evaluation results. This dataset is constructed by extracting 50 samples from each of the different Snowball iterations, 15 iterations in total. As shown in Table~\ref{tab:pscore}, the logic consistency between \BLEC and human evaluation is 0.66 while BLEU, ROUGE, and BLEURT obtain scores below or around zero. This shows that the \BLEC score is capable of testing the logical consistency between logic forms and questions.

\begin{table}[h]
\small
\centering

\begin{tabular}{@{}lll@{}}
\toprule
Metrics & Pearson       & p-value        \\ \midrule
ROUGE \cite{lin-2004-rouge}   & -0.30         & 0.28           \\
BLEU \cite{papineni-etal-2002-bleu}   & -0.24         & 0.40           \\
BLEURT \cite{sellam-etal-2020-bleurt}   & +0.02         & 0.94          \\
\BLEC   & \textbf{+0.66} & \textbf{0.007} \\ \bottomrule
\end{tabular}%

\caption{The correlation with human annotation of BLEU, ROUGE, BLEURT, and our proposed \BLEC.}
\label{tab:pscore}

\end{table}
\begin{table*}[t!]
\resizebox{\textwidth}{!}{%
\begin{tabular}{@{}l|l@{}}
\toprule
Input SQL: & \texttt{SELECT count(*),  \underline{max(Percentage)} FROM country\_language} \\ 
& \hspace{4mm} \texttt{WHERE \underline{LANGUAGE  =  "Spanish"} GROUP BY CountryCode} \\ 
Pre-processed SQL: & ( the number of ( all items ) ) , \underline{( the maximum of ( percentage ) )} that belongs to ( countrylanguage ) , \\
& that have \underline{( ( language ) equal to ( spanish ) )} , grouped by ( countrycode 
)   \\
Label: &  What is the total number of countries where \textbf{Spanish} is spoken by the \textbf{largest} percentage of people? \\ \midrule
BART-base                 & the number and percentage of languages that are \textbf{Spanish} for each country code.      \\
+snowball (iteration = 4) & the number of languages and \textbf{maximum} percentage of languages in each country code?  \\
+multi-tasking            & How many languages does \textbf{Spanish} have in each country code?    \\
\hspace{2mm} +snowball (iteration = 4) &  Show the number and \textbf{maximum} percentage for each country code.     \\ \midrule
BART-large                &  How many languages are there in each country and what is the percentage of the language spoken in that country?    \\
+snowball (iteration = 4) &   Find the number of languages and \textbf{maximum} percentage of \textbf{Spanish} for each country code.\\
+multi-tasking            &  Find the number and percentage of speakers of \textbf{Spanish} in each country code. \\
\hspace{2mm} +snowball (iteration = 4) &  Find the number and \textbf{highest} percentage of speakers of \textbf{Spanish} for each country code  \\ \bottomrule
\end{tabular}%
}
\caption{Example outputs from different models with or w/o performing the MTL and \snowball iteration.}
\label{tab:case_study}
\end{table*}
\subsection{Effectiveness of Snowball Framework}

\begin{table}[t!]
\resizebox{\linewidth}{!}{%
\begin{tabular}{l|lllllllll}
\toprule
                  & \multicolumn{9}{c}{SQL2Text Test Set}                                                               \\ \cline{2-10} 
Metrics           & \multicolumn{6}{c}{\BLEC}                                     & \multicolumn{3}{|c}{Human}  \\ \hline
Snowball           & - & 1 & 2 & 3 & 4 & \multicolumn{1}{l|}{5} & \multicolumn{1}{c}{-} & \multicolumn{1}{l|}{4} & $\kappa$                     \\ \hline
LSTM Seq2Seq      & 22.6 & &  & &  & \multicolumn{1}{l|}{} &  &  &  \\
BART-base         & 76.4 & 78.6 & 78.5 & \textbf{84.1} & 79.7 & \multicolumn{1}{l|}{78.1} & 22 & 45  & 0.69 \\
\hspace{5mm}+MTL        & 89.1 & \textbf{89.5 }& 89.2 & 88.9 & 88.6 & \multicolumn{1}{l|}{88.1} & 66 & 68  & 0.5 \\
BART-large        & 91.8 & 91.3 & \textbf{93.7} & 91.8 & 93.2 & \multicolumn{1}{l|}{93.0} & 75 & 74  & 0.7 \\
\hline
                  & \multicolumn{9}{c}{Logic2Text Test Set}                                                             \\ \cline{2-10} 
Metrics           & \multicolumn{6}{c}{\BLEC}                                     & \multicolumn{3}{|c}{Human}  \\ \hline
Snowball& - & 1 & 2 & 3 & 4 & \multicolumn{1}{l|}{5} & \multicolumn{1}{c}{-} & \multicolumn{1}{l|}{4} & \multicolumn{1}{c}{$\kappa$} \\ \hline
LSTM Seq2Seq        & 41.1 & &  & &  & \multicolumn{1}{l|}{} &  &  &  \\
BART-base         & 87.9 & 86.1 & \textbf{88.6} & 87.4 & 87.7 & \multicolumn{1}{l|}{87.8} & 83 & 85  & 0.48 \\
BART-large        & 86.7 & \textbf{87.8} & 85.2 & 87.1 & 86.0 & \multicolumn{1}{l|}{88.5} & 86 & 78  & 0.48 \\ \bottomrule
\end{tabular}%
}
\caption{The results of \textsc{snowball} generator using \BLEC and human evaluation over different iterations.
}
\label{tab:generator}
\end{table}
\paragraph{Generator}
The experimental results of the generator in our \snowball framework are shown in Table~\ref{tab:generator}.
We found that the \snowball training framework empirically leads to the improvement of the logic consistency. Evaluated by our proposed \BLEC metric, the performance of \texttt{BART-base} generator improves the logic consistency by 10.1\% on SQL2Text and by 0.7\% on Logic2Text. Similarly, the performance of \texttt{BART-large} generator acquires the improvement by 2.1\% on SQL2Text and by 1.2\% on Logic2Text. Under the MTL setting on SQL2Text, the logic faithfulness of the\texttt{BART-base} generator is further enhanced by 16.6\% compared to single-pass training, and is even boosted by 17.1\% by combing with \snowball training.
\begin{figure}[h!]
\includegraphics[width=\linewidth]{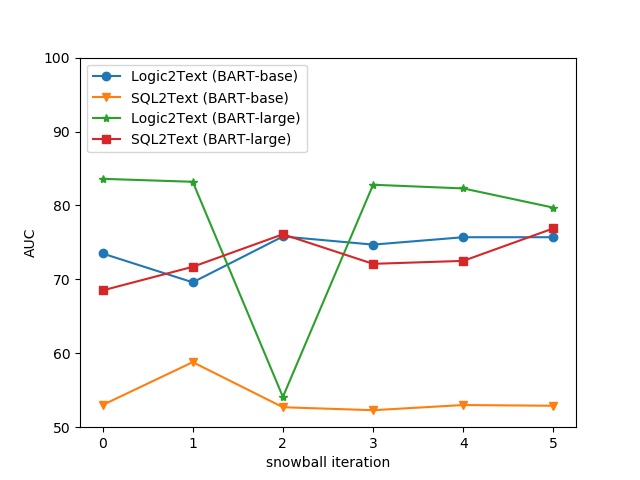}
\caption{The result of \snowball evaluator based on the AUC scores on the test set.
The iteration = 0 means the model is under the regular training procedure as described in Figure~\ref{snowball}.}
\label{fig:auc}
\end{figure}
\paragraph{Evaluator}
The results of the evaluator in our \snowball framework are illustrated in Figure~\ref{fig:auc}. Empirically the snowball framework is more effective to the evaluators base on \texttt{BART-base} than \texttt{BART-large}, this is likely because that the \texttt{BART-large} models have already obtained enough intrinsic knowledge to accurately judge the validness of the [Logical form, Text] pairs. The data augmentation procedure of the \snowball framework may introduce unexpected noise to the evaluators, which may cause a catastrophic reduction in terms of AUC and other metrics. On the other hand, the snowball framework indeed enhances the performance of the evaluator based on relative inferior \texttt{BART-base} by improving the performance on the \textit{SQL2Text} by 10.9\% as well as the \textit{Logic2Text} by 3.1\%. These results indicate that our proposed \snowball framework is most suitable for tasks suffering from both domain data scarcity and the lack of external knowledge.

\subsection{Case Study}

Table \ref{tab:case_study} shows example outputs from our model with different settings. Apparently, in this case, the entity \textbf{Spanish} and the aggregator \textbf{maximum} are the touchstones for evaluating the logic consistency of each model. The prediction from the BART-large based generator trained under both snowball and multi-tasking frameworks simultaneously is the only one that acquires the seamless sentences from the input SQL. Furthermore, we also notice that multi-tasking learning significantly alleviates the artifacts within the generated text. Based on the fact that, compared to the vanilla generators, the generator solely trained with snowball framework would enhance the logic consistency but also increase an unnatural sense to the generated sentences at the same time, we may argue that there is a trade-off between fluency and logic consistency of our purposed snowball framework. The model-level modification may collaboratively enhance the fluency and logic consistency of the NLG, which we would remain for future studies.


\section{Conclusion}
In this paper, we propose \snowball, a neural network-based framework to augment the data alternatively by a generator, and an evaluator. In addition, we propose \BLEC, an automatic evaluation metric that could evaluate the logic consistency between question and logic forms by directional matching. We also formulate two datasets and the experimental results show the effectiveness of the proposed framework. This method is applicable to other Data-to-Text tasks, because domain-specific rules for perturbations can be derived for most structural data with pre-defined structures or grammar.
\section{Ethics Statement}
The datasets we use are built by selecting and processing from two datasets that are open to the public, separately. The data sources we utilize to construct our datasets are Spider and Logic2Text, two complex and cross-domain text-to-SQL datasets. Besides, we use three experts to annotate about 500 data beyond the original dataset. We admit that some biases may still exist in our datasets, even though we have double-checked the data they annotated and the data from the original datasets.

Authors with SQL expertise annotate and verify our datasets through 1) selecting about 500 representative samples from the original dataset, 2) changing the entities of the samples, 3) using three different labels to mark which type of change has been done to the sentences, and 4) double-checking the quality of the data we annotate. 

We conduct several experiments of different settings on our AWS server, with 8 Tesla V100 GPUs, to test the efficiency of our models. To be more specific, our experiments contain two different types. The first type of them is that we train both generator and evaluator using SQL2Text or Logic2Text. The second type of them is that we utilize SQL2Text and Logic2Text to train generators, use one of them to train evaluators in the first epoch, and train the next several epochs with both of them.

\bibliographystyle{acl_natbib}
\bibliography{anthology,acl2021}

\appendix
\clearpage
\newpage

\input{appendix}

\end{document}

%% file: appendix.tex
\section{\BLEC Details}

\BLEC uses bidirectional keyword matching to detect the logic consistency. Table~\ref{tab:blec_rules} shows several sample cases for constructing \BLEC metrics. As shown in the table, the first column shows the type of matching rules. ``Special'' means that these rules are only contained in one of the datasets. Then, the following 3 columns display the tokens in different languages. Using the tokens, the algorithm can detect if the question matches the parse. For instance, given a pair of question and SQL parse, the algorithm could check if ``MAX'' is in SQL parse. If so, it will try to match one of the possible tokens corresponding to ``MAX'', i.e. largest/ greatest, etc., in the question, and vice versa. It is worth noting that, this table only shows a small part of the algorithm, however, all the rules can be classified as one of the three types.

\begin{table}[ht!]
\resizebox{\linewidth}{!}{
\begin{tabular}{|c|l|l|l|}
\hline
Type                      & Spider         & Logic2text      & Natural Language       \\ \hline
Negation                  & NOT            & not\_eq         & not/ none...            \\ \hline
\multirow{4}{*}{Operator} & \textgreater{} & filter\_greater & larger/ more/ greater... \\ \cline{2-4} 
                          & \textless{}    & filter\_smaller & smaller/ less/ fewer...  \\ \cline{2-4} 
                          & MAX            & max             & largest/ greatest...    \\ \cline{2-4} 
                          & COUNT          & count           & total/ how many...      \\ \hline
\multirow{3}{*}{Special}  & ASC            & -               & ascending/ fewest...    \\ \cline{2-4} 
                          & DESC           & -               & descending/ highest...  \\ \cline{2-4} 
                          & -              & most\_str\_eq   & majority/ most...       \\ \hline
\end{tabular}
}
\caption{Sample rules for \BLEC.}
\label{tab:blec_rules}
\end{table}